\documentclass[runningheads]{llncs}
\usepackage{graphicx}
\usepackage{cite}
\usepackage{array}
\usepackage{tabularx,booktabs}

\newcommand{\stimes}{{\times}}

\begin{document}
\title{Garment Recommendation with Memory Augmented Neural Networks}
%
%
\author{Lavinia De Divitiis\orcidID{0000-0002-2485-8218} \and
Federico Becattini\orcidID{0000-0003-2537-2700} \and
Claudio Baecchi\orcidID{0000-0001-8294-4539} \and Alberto Del Bimbo\orcidID{0000-0002-1052-8322}}
\authorrunning{L. De Divitiis et al.}
%
\institute{University of Florence, Italy\\
\email{\{name.surname\}@unifi.it}}
\maketitle              
\begin{abstract}
Fashion plays a pivotal role in society. Combining garments appropriately is essential for people to communicate their personality and style. Also different events require outfits to be thoroughly chosen to comply with underlying social clothing rules. Therefore, combining garments appropriately might not be trivial. The fashion industry has turned this into a massive source of income, relying on complex recommendation systems to retrieve and suggest appropriate clothing items for customers. To perform better recommendations, personalized suggestions can be performed, taking into account user preferences or purchase histories.
In this paper, we propose a garment recommendation system to pair different clothing items, namely tops and bottoms, exploiting a Memory Augmented Neural Network (MANN).
By training a memory writing controller, we are able to store a non-redundant subset of samples, which is then used to retrieve a ranked list of suitable bottoms to complement a given top. In particular, we aim at retrieving a variety of modalities in which a certain garment can be combined.
To refine our recommendations, we then include user preferences via Matrix Factorization.
We experiment on IQON3000, a dataset collected from an online fashion community, reporting state of the art results.

\keywords{Garment Recommendation \and Memory Augmented Neural Networks \and Recommendation Systems.}
\end{abstract}
\section{Introduction}
Recommendation systems are tools that have recently found pervasive use across several application fields. Ranging from social networks to online stores and streaming services, whenever a user has the ability to sift through large databases, machine learning algorithms can come to the aid by suggesting items of possible interest.
A field that is heavily hinged on the effectiveness of recommendation systems is the fashion industry. In fact, revenues for large companies nowadays mostly come from online stores, where users are free to search for desired items. To maximize incomes, it is important for the seller to be able to suggest relevant items, often also exploiting advertisements specifically targeted on user profiles.

Two typical strategies for recommendation are content based and collaborative filtering. The former is based on suggesting similar content to what the user has already purchased in the past. The latter instead exploits a pool of user preferences to infer what a certain individual might like, based on what similar users have purchased.
Both these approaches have limitations. In particular, they can easily end up suggesting several variations of similar items due to their overall popularity or simply because they are close to items owned by the user~\cite{kotkov2020does}. This might even be counterproductive, since it limits the visibility of other items which are relevant yet novel or unexpected. 

In this paper we propose a recommendation system capable of generating a variety of fashion items compatible with a given garment. In particular, we focus on generating different modalities to compose an outfit, rather than suggesting redundant and similar items. To achieve this, we exploit a persistent Memory Augmented Neural Network (MANN), which has proven effective to model diversity~\cite{marchetti2020mantra, marchetti2020multiple}. This kind of models finds its strength in the usage of an external memory, where samples can be explicitly stored at training time and then be read at inference time. Thanks to a trainable memory controller, we are able to store representations of non-redundant outfits, which are then retrieved at inference time to provide recommendations.

For outfit here we refer to a top-bottom pair, e.g. a t-shirt and a skirt. The problem we address is in fact the one of suggesting suitable bottom items, given a top as input. However, differently from prior work~\cite{song2019gp}, we cast the problem as a retrieval task, rather than simply assessing the compatibility between garments.

The main contributions of this paper are the following:

\begin{itemize}
    \item We propose a garment recommendation system based on Memory Augmented Neural Networks. To the best of our knowledge, we are the first to adopt MANNs for garment recommendation.
    \item We train our model to propose a variety of bottoms instead of a set of redundant and almost identical garments, shifting the focus of the retrieval task from single items to garment matching modalities.
    \item We obtain state of the art results on the IQON3000 dataset~\cite{song2019gp}, a collection of outfits collected from an online fashion community.
\end{itemize}

\section{Related Work}

\subsection{Recommendation systems}
In recommendation systems, collaborative filtering methods are often employed to find the relationship between two sets of entities, such as users and products. Among collaborative filtering methods, Matrix Factorization (MF) determines latent features among these sets to find similarities and allows to make predictions on both entities. In \cite{lee2000algorithms} they use a variant of Matrix Factorization called Non-negative Matrix Factorization (NMF), for decomposing multivariate data and study two different algorithms. In \cite{koren2009matrix} the authors employ MF for Netflix movie recommendation to users. They incorporate additional information such as implicit feedback, temporal effects and confidence levels. On a more general scale, \cite{jannach2010recommender} perform an introductory study on collaborative filtering and MF for product recommendation. They also propose a way to measure the effectiveness of recommender systems. In a similar fashion, \cite{ricci2011introduction} illustrate various recommender systems putting particular attention into the difference of the various approaches. Focusing their research on Non-negative MF, \cite{wang2012nonnegative} perform an extensive survey on different methods using NMF, analysing the design principles, characteristics, problems, relationships, and evolution of the algorithms. Finally \cite{bobadilla2013recommender} perform an evolutionary survey on MF and collaborative filtering focusing on the future implications of these methods.

Many recommender systems employ MF for recommending garment matches. In \cite{chang2014multi} they propose a multi-view Non-negative Matrix Factorization (NMF) to solve the problem of clothing matching leveraging multiple types of features. \cite{song2017neurostylist} propose a content-based neural scheme to model the compatibility between fashion items to give meaningful suggestions to the user. In a scenario where multiple items have to be recommended, \cite{wibowo2017matrix} propose several extensions of MF to predict user ratings on garment packaging composed of tops and bottoms. Moreover, they also create a package recommendation dataset to serve their task. Finally, \cite{han2019prototype} propose a way to interpret MF based clothing matching proposals, giving also suggestions on how to interpret them and on how to modify those negative garments to make them appealing to the user. They employ Non-negative MF to suggest the alternatives for each fashion item pair.

\subsection{Garment Recommendation}
In the world of recommendation systems, Garment Recommendation is the task of generating meaningful recommendations of garments based on user interests or on a given input garment. \cite{he2016vbpr} experiment on the \textit{Amazon clothing dataset} to recommend clothes to people based in their interests by using Matrix Factorization techniques. Similarly, in \cite{hu2015collaborative} they give the users suggestions on clothes but considering a set of clothes instead of a single piece. Differently to \cite{he2016vbpr}, they use a functional tensor factorization method to model the interactions between user and fashion items. \cite{song2018neural} leverage the real-world FashionVC dataset to perform complementary clothing matching by integrating deep neural networks and the rich fashion domain knowledge. Recently \cite{bigi2020} leveraged human pose and behaviour to give meaningful suggestions to the user when no prior information about the user is known.

Remaining in the garment recommendation domain, another task that has recently gained interest: Virtual try-on. Virtual try-on complements the recommendation systems with a way of showing the end user how one or more clothes would look if "tried on", simulating the real life act of trying a garment before buying. Here \cite{hauswiesner2011free} leverage  3D cameras to recreate a 3D model of the user and apply on it the selected garment. In \cite{kim2008adoption} they investigate online apparel shoppers’ use of Virtual Try-on to reduce product risks and increase enjoyment in online shopping. The research is based on the electronic Technology Acceptance Model (e-TAM). Finally \cite{zheng2019virtually} introduce a new try-on setting, which enables to change both the clothing item and the person’s pose. They use a pose-guided virtual try-on scheme based on a generative adversarial network (GAN).

\subsection{Memory Augmented Networks}
Memory Augmented Neural Networks (MANN) are a particular declination of Neural Networks that exploit a controller network with an external memory, in which samples can be explicitly stored.
These models have been originally introduced to solve algorithmic tasks~\cite{graves2014neural, santoro2016meta, sukhbaatar2015end, weston2014memory}, however several applications of MANNs have been proposed in literature~\cite{yang2018learning, ma2018visual, pernici2018memory, pernici2020self, marchetti2020multiple}.
The first work to propose a model equipped with an external memory has been Neural Turing Machines (NTM)~\cite{graves2014neural}. The authors exploited the MANN as a working memory to solve data manipulation tasks, such as sorting or copying.

The problem of online learning has also been addressed with MANNs. Pernici \textit{et al.}~\cite{pernici2020self} rely on an external memory to incrementally store people identities for re-identification purposes. In~\cite{yang2018learning}, Memory Augmented Networks have been shown to aid object tracking, by offering the ability to store templates of the tracked objects. Also incremental classifiers have been implemented~\cite{rebuffi2017icarl}, by adding classes in memory, incrementally.

Several works have found MANNs beneficial for Question Answering~\cite{weston2014memory, sukhbaatar2015end} or Visual Question Answering~\cite{kumar2016ask, ma2018visual} tasks. In these works, some attention mechanism is usually exploited to guide the answering process by sequentially attending to different memory locations or to assign a different importance to common and uncommon question answer pairs.

Most of these approaches, however, rely on episodic memories. This means that the model learns how to manipulate data, rather than learning to build a persistent collection of samples.
Recently, a model for trajectory forecasting has been proposed by Marchetti \textit{et al.}~\cite{marchetti2020mantra, marchetti2020multiple}, where the model learns to store non redundant samples relying on a specifically tailored writing controller. In this work, we rely on a similar strategy, by adding in memory a collection of garments. We then use these to retrieve different modalities to create an outfit by matching clothing items.
Differently from~\cite{marchetti2020mantra}, we make use of a convolutional model to store image embeddings in memory and we cast the model to solve a recommendation task by adding user information via Matrix Factorization.



\section{Garment Recommendation with MANNs}
We focus on the problem of garment recommendation, i.e. the task of retrieving a ranked list of bottoms that can be combined with a a given top. Tops and bottoms are clothing items that can be worn together, defining an outfit.
We refer to the set of top and bottom images respectively as $\mathcal{T}=\{ t_0, t_1, ... t_{N_T} \}$ and $\mathcal{B}=\{ b_0, b_1, ... b_{N_B} \}$, where $N_T$ and $N_B$ are the total number of garments in the two sets.
Each user $u_k \in \mathcal{U}=\{ u_0, u_1, ... u_{N_U} \}$ is associated with a collection of outfits, i.e. top-bottom pairs which have been rated as positive matches. We indicate this outfit set for user $u_k$ as $\mathcal{O}_k = \{ (t_i, t_j) \}_{ij}$, where $i$ and $j$ act as indexes selecting items in $\mathcal{T}$ and $\mathcal{B}$.

The goal of the garment recommendation task can therefore be formulated as the task of retrieving the bottom $t_j$ given a user $u_k$ and a top $t_i$ belonging to $\mathcal{O}_k$.

In the following, we first present our Memory Augmented Neural Network based method for establishing matches according to general style preferences and we then outline our recommendation refinement strategy, which includes personal user preferences via Matrix Factorization.

\subsection{Feature Representation}
\label{sec:autoencoder}
In order to let our model manipulate tops and bottoms and store them into its external memory, we first need to devise a way to extract compact representations.
To this end, we train an autoencoder that jointly learns to reconstruct top and bottom items forming an outfit.

\begin{figure}[!t]
	\centering
	\includegraphics[width=\columnwidth]{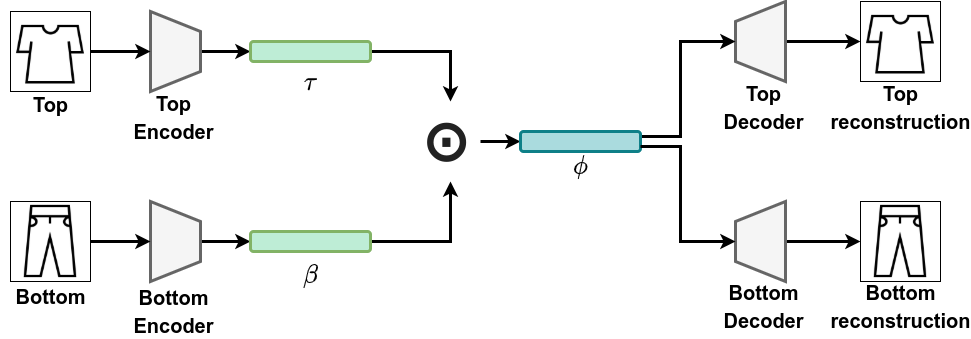}
	\caption{Top and bottom images are separately encoded using convolutional encoders. The resulting features $\tau$ and $\beta$ are then combined through a scalar product into a common representation $\phi$. Finally, $\phi$ is decoded into two separate reconstructions with deconvolutional layers to obtain an approximation of the inputs.}
	\label{fig:autoencoder}
\end{figure}

Separate convolutional encoders project top and bottom images into a common latent space. We denote the top encoding as $\tau$ and the bottom encoding as $\beta$. The two features are then blended together with a scalar product into a joint representation $\phi$ and are then separately reconstructed with two deconvolutional decoders. Fig.~\ref{fig:autoencoder} depicts the structure of the model. In our experiments we observed that training the top and bottom autoencoders jointly led to a better generalization, reducing overfitting.

The benefits of using an autoencoder are twofold: on the one hand, we are able to obtain a compact representation of garments without any further annotation cost; on the other hand, we are also training a decoding function which will be useful for extracting bottoms from memory.


We use the same architecture for the two pairs of encoders and decoders. As for the encoder, we perform three $3\stimes3$ 2D convolutions with $8, 16$ and $32$ filters respectively. All three convolutions have a padding applied of $1$ pixel. We then apply $3\stimes3$ max pooling with a stride of $3$. The relu activation function is applied after each convolution.
On the decoder side, we employ three 2D transposed convolutions to attempt to reconstruct the input. Specifically, we set the number of channels to $16, 8$ and $3$ with a kernel size of $5\stimes5, 8\stimes8$ and $3\stimes3$ respectively. As for padding we apply $4, 2$ and $3$ pixels respectively, with an extra pixel of output padding in the first transposed 2D convolution. The relu activation function is applied after each transposed convolution.
Finally to each encoder-decoder pair output we apply the sigmoid function to output values in the range $[0, 1]$.

\subsection{Model}
To find suitable bottom garments to be paired with a given top, we adopt a model based on a Memory Augmented Neural Network. The model is trained to identify a non-redundant subset of outfits to be stored in memory, which can then be used to produce recommendations. The memory acts as a learnable interface between user queries and the underlying database. Since our aim is to suggest multiple outfit modalities, rather than retrieving all suitable instances in the dataset, we train the model to store only samples that can be used to suggest a variety of bottoms.

Given an outfit composed of a top-bottom pair ($t$, $b$), we want our model to output a ranked list of bottoms, where at least one is sufficiently close to $b$, when $t$ is fed as input.
To this end, when the top $t$ is fed to the model, we encode it into a feature representation $\tau$, using the encoding from Sec.~\ref{sec:autoencoder}. This compact representation of the input is then used as key to access the external memory. Memory access is performed via cosine similarity between the key and the top component of each stored outfit.
This generates a distribution of similarity scores $s_i$ for each element in memory:
\begin{equation}
s_i = \frac{\tau \cdot \tau_i}{\|\tau\| \cdot \|\tau_i\|} ~~~~~ for ~~ i=0,...,|M|-1
\end{equation}
where $|M|$ is the memory size.

\begin{figure}[!t]
	\centering
	\includegraphics[width=\columnwidth]{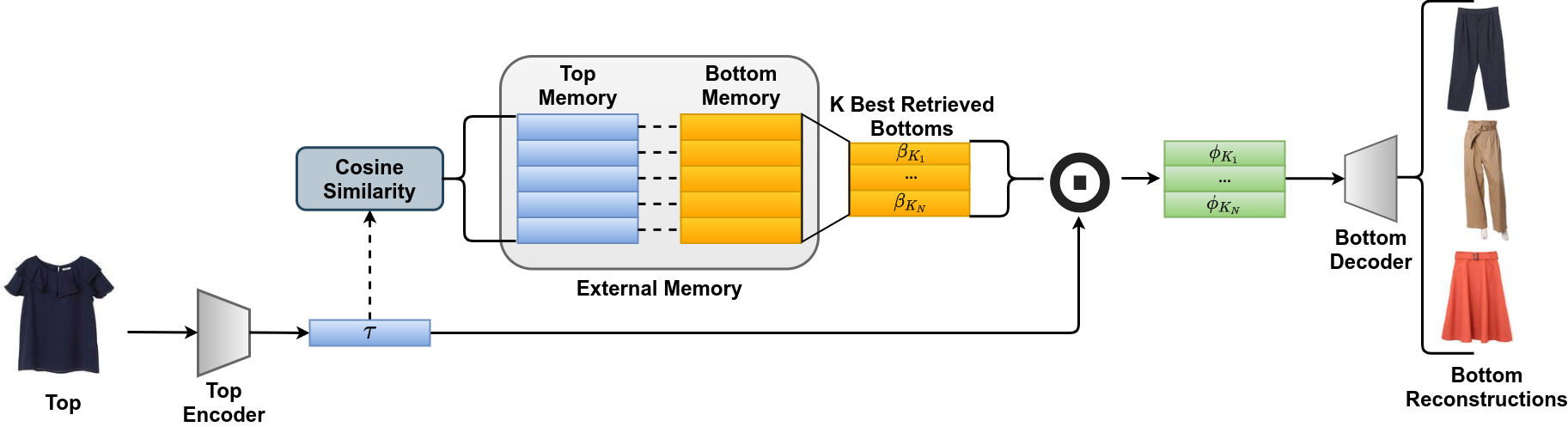}
	\caption{Given an input image depicting a top, we encode it into a latent representation $\tau$ and use it to retrieve samples from memory via cosine similarity. The bottom part of the best samples are then combined with $\tau$ and decoded into bottom reconstructions.}
	\label{fig:model}
\end{figure}

To generate a ranked set of recommendations, we take the $K_N$ samples that give the highest scores. For each outfit retrieved from memory, we take its bottom $\beta_{K_i}$ and we combine it with $\tau$ via dot product, to obtain a joint representation $\phi_{K_i}$. Each feature is then fed to the decoder part of the autoencoder (see Section~\ref{sec:autoencoder}) to reconstruct different bottom images. Each reconstruction is performed in parallel, independently from the others. Fig.~\ref{fig:model} shows the inference process of our model.

To guarantee diversity, as well as satisfactory reconstructions, we train a writing controller that decides what samples to insert into memory.
At training time, a top is fed to the network and $K_N$ bottoms are proposed as output, as described above. The task of garment recommendation is inherently multimodal, i.e. multiple outputs might be considered correct given a single input. In other terms, a user might consider multiple bottoms to be a good match with a top to create different outfits.

Since we only have one ground truth bottom $t^*$, we cannot explicitly add supervision on all the generated outputs. 
To overcome this limitation, we borrow a concept from the trajectory forecasting literature, where the multimodality of the task plays an even more central role. In fact, we derive our controller loss from the Variety Loss~\cite{gupta2018social} to train our network.
The Variety Loss, often referred to as Best-of-K loss, encourages the generation of diverse samples $\hat{t}_k$ by simply backpropagating the error only for the best prediction, leaving the other predictions free to explore the output space:

\begin{equation}
    \mathcal{L}_{Variety} = \min_{k} \| \hat{t}_k - t^{*} \|_2
\end{equation}

Our memory writing controller, i.e. a trainable feedforward network that outputs a writing probability $P_w$ for a given sample, takes as input the error value given by the Variety Loss itself. Based on the best reconstruction error, in fact, it establishes whether the current sample should be written to memory. The rationale behind this idea is that a sample only needs to be written if the model is not capable of generating a sufficiently close reconstruction of the ground truth.

We therefore optimize the controller loss $\mathcal{L}_{Controller}$ defined as follows:

\begin{equation}
    \mathcal{L}_{Controller} = \mathcal{L}_{Variety}(1 - P_w) + (1 - \mathcal{L}_{Variety})P_w
\end{equation}

A similar loss has been adopted in~\cite{marchetti2020mantra} to fill a persistent memory with non-redundant samples. The controller generates a writing probability in $[0, 1]$ and, since we use normalized images and an MSE error function for the Variety Loss, $\mathcal{L}_{Variety}$ also yields values in $[0, 1]$. When the reconstruction generates a high error, then $\mathcal{L}_{Variety}$ will be close to 1 and the sample should be written to improve the recommendation capabilities of the network. In this case the loss will optimize the controller to output a high writing probability. On the contrary, when the reconstruction error is low, the controller will be optimized to discard the observed outfit.

When a sample gets written in memory, we store the pair of encodings for the top and bottom. At training time, we start from a randomly initialized memory with $K_N$ samples, to be able to always suggest a sufficient number of elements. We empty the memory after each epoch and populate the final memory after the controller has been trained by iterating once again on the training set.


\subsection{General recommendations and user preferences}
The model presented so far is capable or retrieving bottoms that can be matched with a given top. These matches, however, follow the general fashion criteria underlying the data distribution. Effective recommendation systems typically strive at suggesting items that are targeted for specific users, based on some past information, such as purchase history, or similarity with other users.

A common approach to generate targeted recommendations is Matrix Factorization~\cite{wang2012nonnegative, bobadilla2013recommender}. In our garment recommendation system, we first retrieve a set of bottoms with our Memory Augmented Neural Network. These bottoms represent different modalities to compose an outfit based on the given top. Although the MANN is capable of providing an ordered set according to top similarity with stored samples, this ranking does not take into account the user.

We refine our recommendations by re-ranking the set of bottoms via Matrix Factorization, which generates compatibility scores between the user and each bottom. Matrix Factorization projects the user and bottom IDs into a common space using trainable encoding functions and combines them together with a dot product. For both the user and the proposed bottoms, we also learn a bias to shift the decision. The final compatibility score, with values in $[0, 1]$, is obtained through a sigmoid activation.

To be able to work with known garments, i.e. belonging to a given database collection, we retrieve the most similar bottoms in the training set to the bottoms reconstructed by the decoder of the MANN. This allows us to learn a meaningful projection for bottoms, based on all user preferences.

\section{Experiments}
In the following we first provide some details on the dataset and metrics used to evaluate our model and then we report results, comparing the proposed approach with the state of the art.

\subsection{Dataset and Metrics}
\label{sec:dataset}
For our experiments, we use the IQON3000 dataset~\cite{song2019gp}. This dataset contains a collection of 308,747 outfits, handpicked by 3,568 different users. Each outfit comprises garments belonging to 6 categories: \textit{coat}, \textit{top}, \textit{bottom}, \textit{one piece}, \textit{shoes} and \textit{accessories}. In total the dataset has 672,335 fashion items. All items are labeled with attributes such as \textit{color}, \textit{category}, \textit{price} and \textit{description}.
Similarly to prior work~\cite{song2019gp}, in this paper we focus solely on top and bottom garments and we always consider tops to be given as inputs in order to find matching bottoms. The proposed model, however, could be easily adapted to the inverse problem by simply swapping tops and bottoms, without any further modification.

The dataset was first introduced to address the task of personalized garment matching. This was originally engineered to simply assess whether a model was capable of assigning a higher score to known positive matches compared to random ones. Here we extend this task, aiming at obtaining a ranked list of recommendations among all possible garments. However, we believe that an effective recommendation system should propose a variety of different outputs instead of suggesting several small variations of the same garment.

To this end, instead of simply comparing the ID of the retrieved bottoms, we compare their categories and colors. Proposing bottoms of different colors and categories, to be paired with the same top, means that we are proposing diverse modalities to combine the given top and generate an outfit. In the dataset there are 62 bottom categories, 12 different colors and a total of 742 combinations of the two.

We evaluate our approach by measuring the Accuracy@K, i.e. the fraction of samples for which the model suggests at least one bottom sharing the same category and/or color of the ground truth, among the first K retrieved items.
In addition, we report the quality of the ranking by measuring the mean Average Precision (mAP). This is of interest especially when considering user personal preferences.

\newcolumntype{?}{!{\vrule width 1pt}}
\newcolumntype{Y}{>{\centering\arraybackslash}X}
\begin{table}[t]
\begin{tabularx}{\textwidth}{ c? Y|Y?Y|Y?YY }
   & \multicolumn{2}{c?}{Category $\times$ Color} & \multicolumn{2}{c?}{Category} & \multicolumn{2}{c}{Color} \\ \hline
 Num items & Ours             & GP-BPR            & Ours         & GP-BPR        & Ours       & GP-BPR       \\ \hline
5               & \textbf{30}               & 27                & \textbf{81}           & 75            & \textbf{58}         & 55           \\
10              & \textbf{45}               & 44                & \textbf{89}           & 84            & \textbf{73}         & 70           \\
20              & \textbf{59}               & 56                & \textbf{93}           & 90            & \textbf{85}         & 81           \\
30              & \textbf{67}               & 63                & \textbf{95}           & 92            & \textbf{91}         & 87           \\
40              & \textbf{71}               & 69                & \textbf{96}           & 93            & \textbf{94}         & 90           \\
50              & \textbf{75}               & 72                & \textbf{96}           & 94            & \textbf{96}         & 92           \\
60              & \textbf{78}               & 74                & \textbf{97}           & 94            & \textbf{97}         & 92           \\
\end{tabularx}
	\caption{Comparison between our model and GP-BPR~\cite{song2019gp}. An item is considered correct if it shares Category and/or Color with the ground truth.}
	\label{tab:acc}
\end{table}

\subsection{Results}
Here we discuss the results obtained by our model. First of all, we evaluate the accuracy of the model by considering colors and categories, as explained in Sec~\ref{sec:dataset}.
We measure the accuracy varying the number of retrieved items, as shown in Tab.~\ref{tab:acc}. Despite the number of categories is approximately five times higher than the number of colors, the system immediately obtains an accuracy above 80\%, even with just 5 retrieved items. Considering colors, instead, the accuracy starts from around 60\%. We have attributed this to the fact that color matching, despite following precise social rules, allows some flexibility with similar tonalities.

As expected, when considering both color and category, the accuracy drops considerably. Interestingly, in all cases, we are able to obtain a higher accuracy than the state of the art. In particular, we report results obtained by GP-BPR~\cite{song2019gp}. The results are obtained with the pre-trained models released by the authors by generating all compatibility scores between the pair of current top and user and all the bottoms in the dataset. We then take the highest ranking garments to generate the recommendations.

\begin{figure}[!t]
	\centering
	\includegraphics[width=\columnwidth]{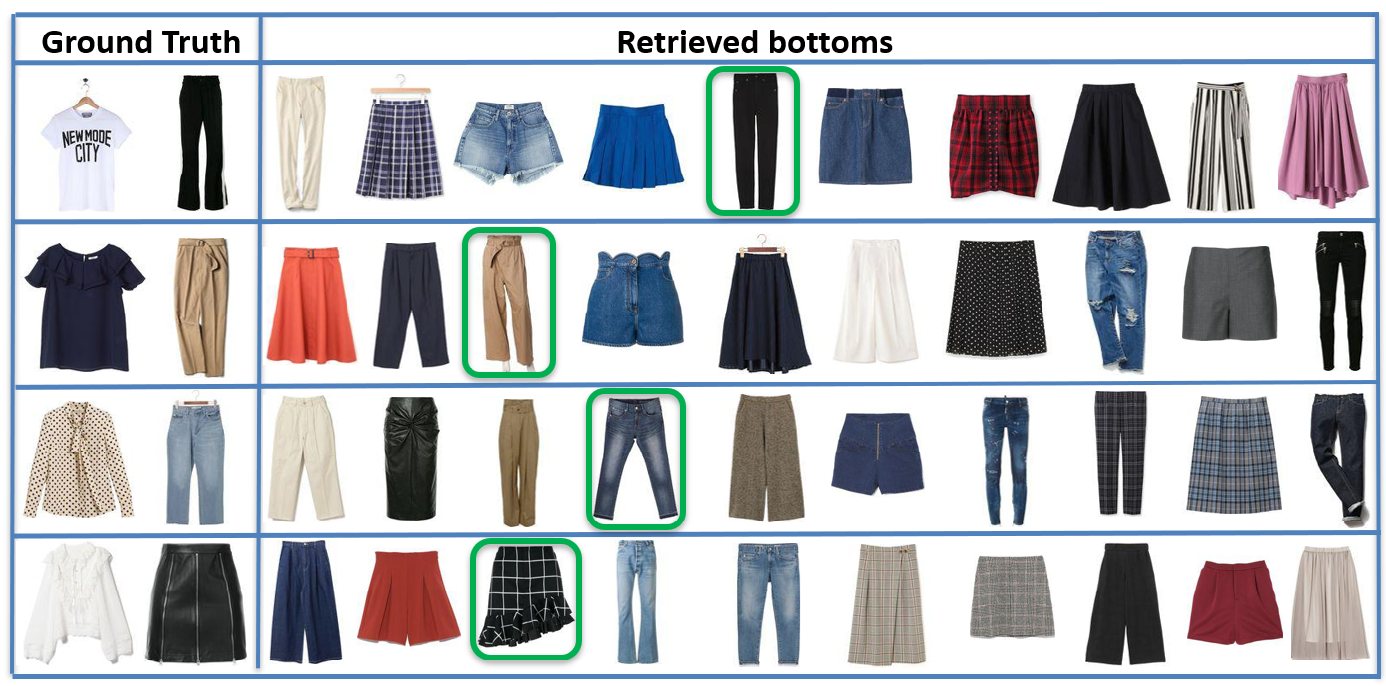}
	\caption{Qualitative results generated by our proposed model. The top is given as input and a ranked set of bottom is retrieved. It can be observed that bottoms similar to the ground truth appear among the first recommendations.}
	\label{fig:qualitative}
\end{figure}

Despite GP-BPR explicitly considers the compatibility with each bottom in the dataset, our approach is still able to yield a higher accuracy relying only on the subset of garments stored in memory.
Fig.~\ref{fig:qualitative} shows some qualitative samples of bottoms retrieved from memory. It can be seen how there is a high variability in the suggested outfits and that a similar bottom to the ground truth is found in the first positions of the rankings.

\begin{figure}[!t]
	\centering
	\includegraphics[width=.8\columnwidth]{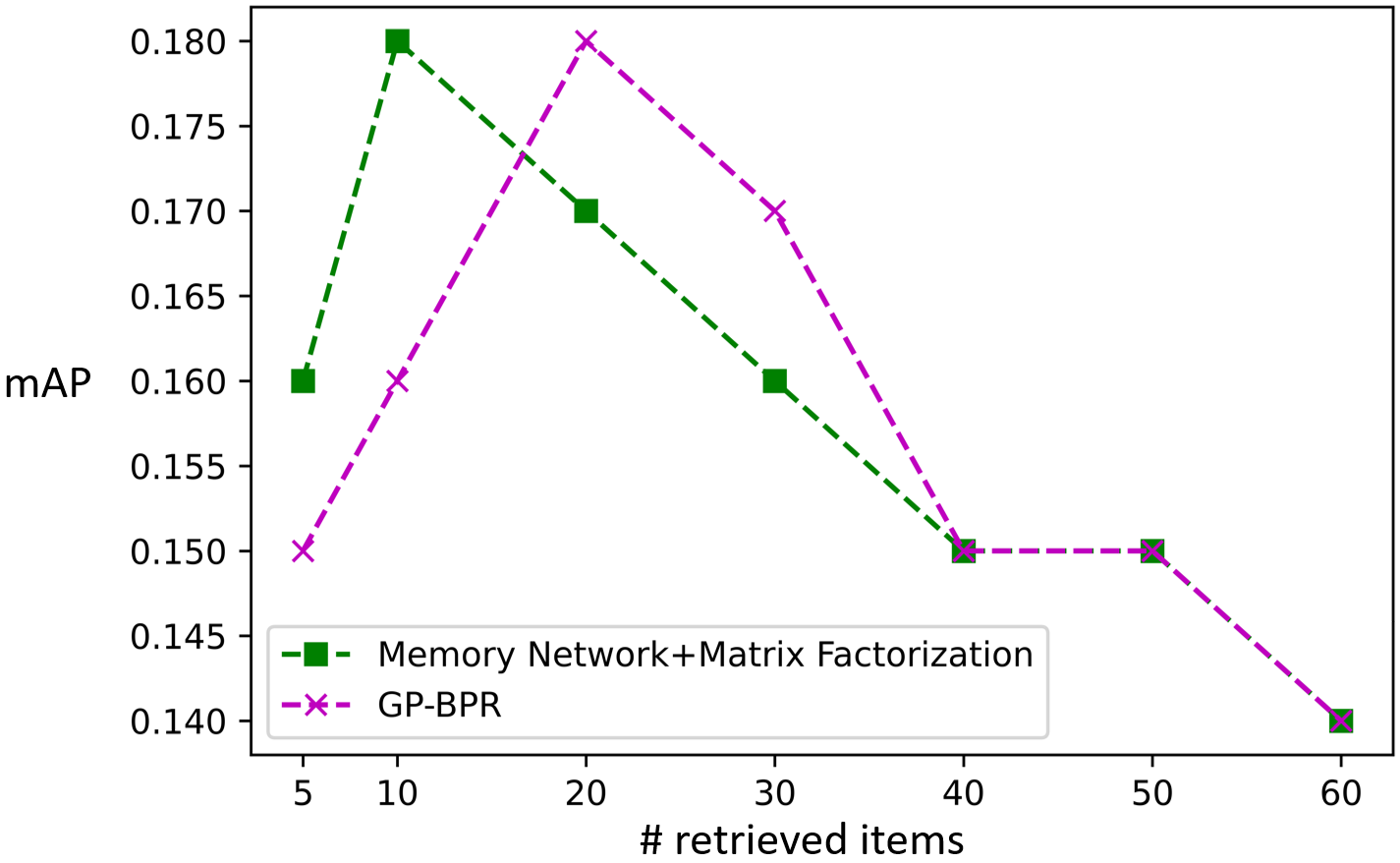}
	\caption{Mean Average Precision of the proposed approach using Memory Network and Matrix Factorization. As reference, we report the results obtained by GP-BPR~\cite{song2019gp}. Interestingly, with a low number of retrieved items, our method obtained a higher mAP compared to GP-BPR.}
	\label{fig:map}
\end{figure}

We then measure the mean Average Precision obtained by our system. Fig~\ref{fig:map} shows the curve, varying the number of retrieved garments. The combination of the Memory Network with Matrix Factorization yields to better results than GP-BPR~\cite{song2019gp} for a low number of recommended items, which is the scenario with the most relevant applicability.

\begin{figure}[!t]
	\centering
	\includegraphics[width=.8\columnwidth]{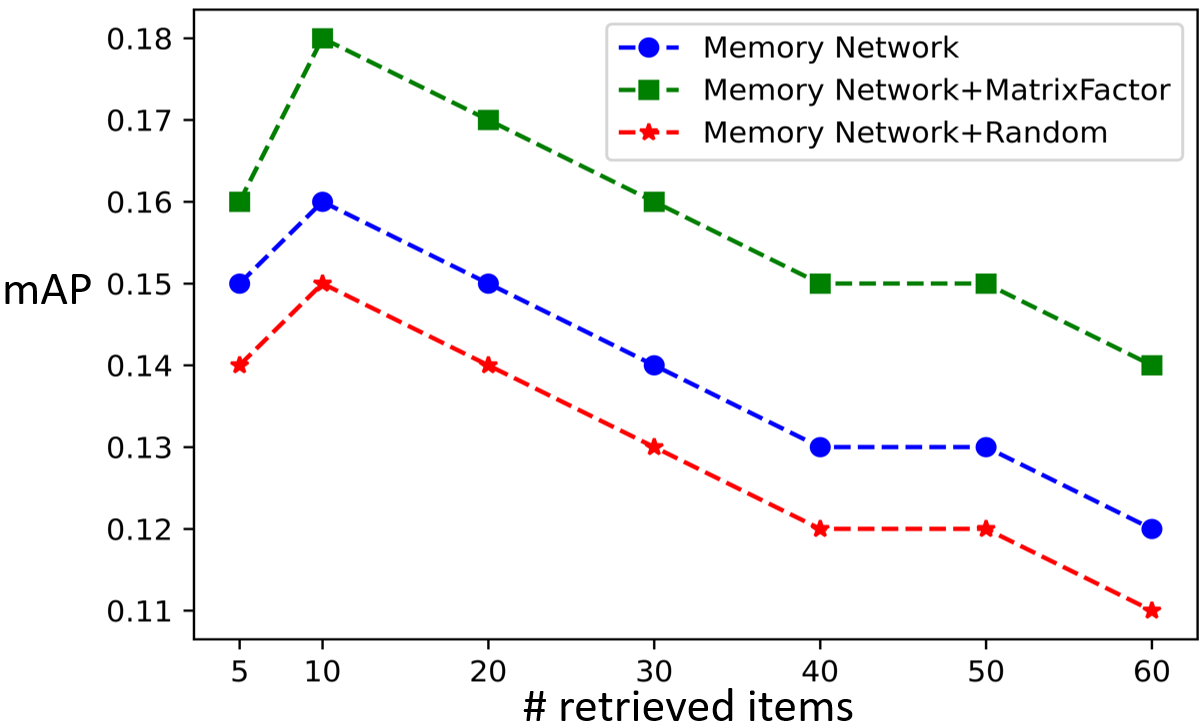}
	\caption{Impact of reranking on the garments retrieved from the Memory Network. Matrix Factorization provides an improvement including user preferences.}
	\label{fig:reranking}
\end{figure}

\begin{figure}[!t]
	\centering
	\includegraphics[width=.8\columnwidth]{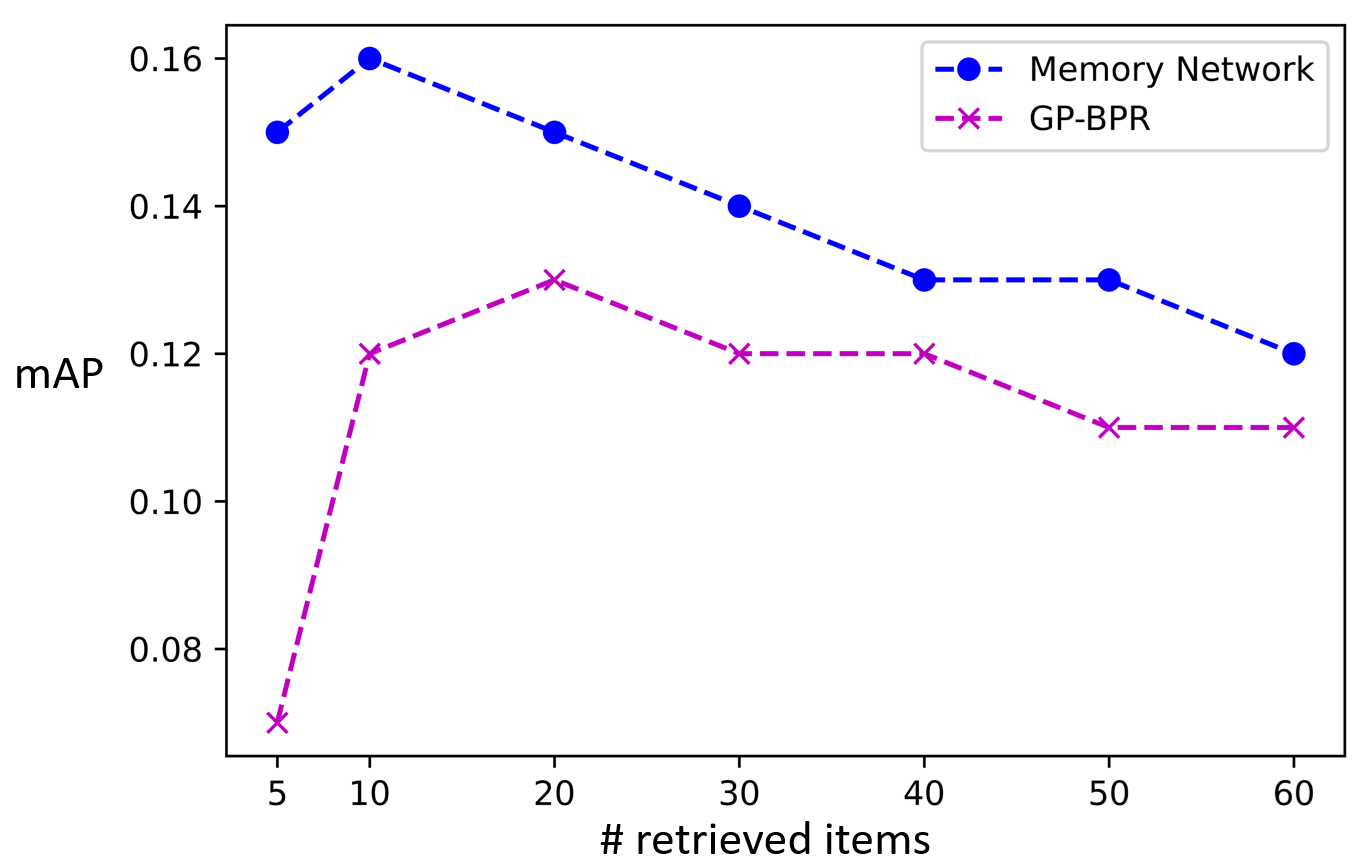}
	\caption{Comparison of our proposed method against GP-BPR~\cite{song2019gp} without including user information.}
	\label{fig:map_no_user}
\end{figure}

Finally we show in Fig.~\ref{fig:reranking} and Fig.~\ref{fig:map_no_user} the effect of removing user information in the recommendation pipeline. In this way, we obtain only recommendations that follow a common sense, rather than personalized suggestions. Again, we compare our results against~\cite{song2019gp}, removing the user information also for this model for a fair comparison. It can be seen in Fig.~\ref{fig:reranking} how our method reports a 2\% drop, while on the other hand, GP-BPR suffers from a much higher loss (Fig.~\ref{fig:map_no_user}). This suggests that our memory network is capable of providing a satisfactory set of garments even without considering the user.
At the same time, we verify that Matrix Factorization is capable of reranking the garments effectively and that overall the Memory Network, with and without Matrix Factorization, are capable or providing better results compared to a Random reranking baseline.

\section{Conclusions}
In this paper we have presented a garment recommendation system based on a Memory Agumented Neural Network. The model relies on an external memory, which is filled by a trainable controller to store non redundant samples that capture multiple modalities to match top and bottom garments. In this way, the system is capable of retrieving and suggesting a variety of garments instead of proposing small variations of the same outfit. We have shown how this reflects on improved recommendation capabilities compared to the state of the art.

\section*{Acknowledgements}
This work was partially supported by the Italian MIUR within PRIN 2017, Project Grant 20172BH297: I-MALL - improving the customer experience in stores by intelligent computer vision.

%
%
%
\bibliographystyle{splncs04}
\bibliography{bib.bib}

\begin{thebibliography}{10}
\providecommand{\url}[1]{\texttt{#1}}
\providecommand{\urlprefix}{URL }
\providecommand{\doi}[1]{https://doi.org/#1}

\bibitem{bigi2020}
Bigi, W., Baecchi, C., Del~Bimbo, A.: Automatic interest recognition from
  posture and behaviour. In: Proceedings of the 28th ACM International
  Conference on Multimedia. p. 2472–2480. MM '20, Association for Computing
  Machinery, New York, NY, USA (2020). \doi{10.1145/3394171.3413530},
  \url{https://doi.org/10.1145/3394171.3413530}

\bibitem{bobadilla2013recommender}
Bobadilla, J., Ortega, F., Hernando, A., Guti{\'e}rrez, A.: Recommender systems
  survey. Knowledge-based systems  \textbf{46},  109--132 (2013)

\bibitem{chang2014multi}
Chang, W.Y., Wei, C.P., Wang, Y.C.F.: Multi-view nonnegative matrix
  factorization for clothing image characterization. In: 2014 22nd
  International Conference on Pattern Recognition. pp. 1272--1277. IEEE (2014)

\bibitem{graves2014neural}
Graves, A., Wayne, G., Danihelka, I.: Neural turing machines. arXiv preprint
  arXiv:1410.5401  (2014)

\bibitem{gupta2018social}
Gupta, A., Johnson, J., Fei-Fei, L., Savarese, S., Alahi, A.: Social gan:
  Socially acceptable trajectories with generative adversarial networks. In:
  Proceedings of the IEEE Conference on Computer Vision and Pattern
  Recognition. pp. 2255--2264 (2018)

\bibitem{han2019prototype}
Han, X., Song, X., Yin, J., Wang, Y., Nie, L.: Prototype-guided attribute-wise
  interpretable scheme for clothing matching. In: Proceedings of the 42nd
  International ACM SIGIR Conference on Research and Development in Information
  Retrieval. pp. 785--794 (2019)

\bibitem{hauswiesner2011free}
Hauswiesner, S., Straka, M., Reitmayr, G.: Free viewpoint virtual try-on with
  commodity depth cameras. In: Proceedings of the 10th International Conference
  on Virtual Reality Continuum and Its Applications in Industry. pp. 23--30
  (2011)

\bibitem{he2016vbpr}
He, R., McAuley, J.: Vbpr: visual bayesian personalized ranking from implicit
  feedback. In: Thirtieth AAAI Conference on Artificial Intelligence (2016)

\bibitem{hu2015collaborative}
Hu, Y., Yi, X., Davis, L.S.: Collaborative fashion recommendation: A functional
  tensor factorization approach. In: Proceedings of the 23rd ACM international
  conference on Multimedia. pp. 129--138 (2015)

\bibitem{jannach2010recommender}
Jannach, D., Zanker, M., Felfernig, A., Friedrich, G.: Recommender systems: an
  introduction. Cambridge University Press (2010)

\bibitem{kim2008adoption}
Kim, J., Forsythe, S.: Adoption of virtual try-on technology for online apparel
  shopping. Journal of Interactive Marketing  \textbf{22}(2),  45--59 (2008)

\bibitem{koren2009matrix}
Koren, Y., Bell, R., Volinsky, C.: Matrix factorization techniques for
  recommender systems. Computer  \textbf{42}(8),  30--37 (2009)

\bibitem{kotkov2020does}
Kotkov, D., Veijalainen, J., Wang, S.: How does serendipity affect diversity in
  recommender systems? a serendipity-oriented greedy algorithm. Computing
  \textbf{102}(2),  393--411 (2020)

\bibitem{kumar2016ask}
Kumar, A., Irsoy, O., Ondruska, P., Iyyer, M., Bradbury, J., Gulrajani, I.,
  Zhong, V., Paulus, R., Socher, R.: Ask me anything: Dynamic memory networks
  for natural language processing. In: International conference on machine
  learning. pp. 1378--1387 (2016)

\bibitem{lee2000algorithms}
Lee, D., Seung, H.S.: Algorithms for non-negative matrix factorization.
  Advances in neural information processing systems  \textbf{13},  556--562
  (2000)

\bibitem{ma2018visual}
Ma, C., Shen, C., Dick, A., Wu, Q., Wang, P., van~den Hengel, A., Reid, I.:
  Visual question answering with memory-augmented networks. In: Proceedings of
  the IEEE Conference on Computer Vision and Pattern Recognition. pp.
  6975--6984 (2018)

\bibitem{marchetti2020mantra}
Marchetti, F., Becattini, F., Seidenari, L., Bimbo, A.D.: Mantra: Memory
  augmented networks for multiple trajectory prediction. In: Proceedings of the
  IEEE/CVF Conference on Computer Vision and Pattern Recognition. pp.
  7143--7152 (2020)

\bibitem{marchetti2020multiple}
Marchetti, F., Becattini, F., Seidenari, L., Del~Bimbo, A.: Multiple trajectory
  prediction of moving agents with memory augmented networks. IEEE Transactions
  on Pattern Analysis and Machine Intelligence  (2020)

\bibitem{pernici2018memory}
Pernici, F., Bartoli, F., Bruni, M., Del~Bimbo, A.: Memory based online
  learning of deep representations from video streams. In: Proceedings of the
  IEEE Conference on Computer Vision and Pattern Recognition. pp. 2324--2334
  (2018)

\bibitem{pernici2020self}
Pernici, F., Bruni, M., Del~Bimbo, A.: Self-supervised on-line cumulative
  learning from video streams. Computer Vision and Image Understanding p.
  102983 (2020)

\bibitem{rebuffi2017icarl}
Rebuffi, S.A., Kolesnikov, A., Sperl, G., Lampert, C.H.: icarl: Incremental
  classifier and representation learning. In: Proceedings of the IEEE
  conference on Computer Vision and Pattern Recognition. pp. 2001--2010 (2017)

\bibitem{ricci2011introduction}
Ricci, F., Rokach, L., Shapira, B.: Introduction to recommender systems
  handbook. In: Recommender systems handbook, pp. 1--35. Springer (2011)

\bibitem{santoro2016meta}
Santoro, A., Bartunov, S., Botvinick, M., Wierstra, D., Lillicrap, T.:
  Meta-learning with memory-augmented neural networks. In: International
  conference on machine learning. pp. 1842--1850 (2016)

\bibitem{song2018neural}
Song, X., Feng, F., Han, X., Yang, X., Liu, W., Nie, L.: Neural compatibility
  modeling with attentive knowledge distillation. In: The 41st International
  ACM SIGIR Conference on Research \& Development in Information Retrieval. pp.
  5--14 (2018)

\bibitem{song2017neurostylist}
Song, X., Feng, F., Liu, J., Li, Z., Nie, L., Ma, J.: Neurostylist: Neural
  compatibility modeling for clothing matching. In: Proceedings of the 25th ACM
  international conference on Multimedia. pp. 753--761 (2017)

\bibitem{song2019gp}
Song, X., Han, X., Li, Y., Chen, J., Xu, X.S., Nie, L.: Gp-bpr: Personalized
  compatibility modeling for clothing matching. In: Proceedings of the 27th ACM
  International Conference on Multimedia. pp. 320--328 (2019)

\bibitem{sukhbaatar2015end}
Sukhbaatar, S., Weston, J., Fergus, R., et~al.: End-to-end memory networks. In:
  Advances in neural information processing systems. pp. 2440--2448 (2015)

\bibitem{wang2012nonnegative}
Wang, Y.X., Zhang, Y.J.: Nonnegative matrix factorization: A comprehensive
  review. IEEE Transactions on Knowledge and Data Engineering  \textbf{25}(6),
  1336--1353 (2012)

\bibitem{weston2014memory}
Weston, J., Chopra, S., Bordes, A.: Memory networks. arXiv preprint
  arXiv:1410.3916  (2014)

\bibitem{wibowo2017matrix}
Wibowo, A.T., Siddharthan, A., Lin, C., Masthoff, J.: Matrix factorization for
  package recommendations. In: Proceedings of the RecSys 2017 Workshop on
  Recommendation in Complex Scenarios (ComplexRec 2017). CEUR-WS (2017)

\bibitem{yang2018learning}
Yang, T., Chan, A.B.: Learning dynamic memory networks for object tracking. In:
  Proceedings of the European Conference on Computer Vision (ECCV). pp.
  152--167 (2018)

\bibitem{zheng2019virtually}
Zheng, N., Song, X., Chen, Z., Hu, L., Cao, D., Nie, L.: Virtually trying on
  new clothing with arbitrary poses. In: Proceedings of the 27th ACM
  International Conference on Multimedia. pp. 266--274 (2019)

\end{thebibliography}

\end{document}